\pdfoutput=1

\documentclass[11pt]{article}

\usepackage{emnlp2023}

\usepackage{times}
\usepackage{latexsym}

\usepackage[T1]{fontenc}

\usepackage[utf8]{inputenc}

\usepackage{microtype}

\usepackage{inconsolata}

\usepackage{graphicx}
\usepackage{amsmath}
\usepackage{enumitem}
\usepackage{booktabs}
\usepackage{tabularx}
\usepackage{multirow, makecell}
\usepackage[normalem]{ulem}
\usepackage{CJKutf8}
\usepackage{color}

\definecolor{darkred}{rgb}{0.5,0.0,0.0}
\definecolor{darkgreen}{rgb}{0.0,0.6,0.0}

\useunder{\uline}{\ul}{}
\newcommand{\dr}[1]{\textcolor{darkred}{\bf\fontsize{9}{12} \selectfont \,(#1\%)}}
\newcommand{\drs}[1]{\textcolor{darkred}{\bf\fontsize{9}{12} \selectfont \,\phantom{0}(#1\%)}}

\newcommand{\dgp}[1]{\textcolor{darkgreen}{\bf\fontsize{9}{12} \selectfont \,\phantom{-}(#1\%)}}
\newcommand{\dgsp}[1]{\textcolor{darkgreen}{\bf\fontsize{9}{12} \selectfont \,\phantom{-}\phantom{0}(#1\%)}}

\newcommand{\prism}{Prism}
\newcommand{\ftprism}{Prism+FT}

\title{Trained MT Metrics Learn to Cope with Machine-translated References}

 \author{
    {\bf Jannis Vamvas$^2$\thanks{~~Work done during an internship at Amazon.} \hspace{2mm}}
    {\bf Tobias Domhan$^1$ \hspace{2mm}}
    {\bf Sony Trenous$^1$ \hspace{2mm}} \\
    {\bf Rico Sennrich$^2$ \hspace{2mm}}
    {\bf Eva Hasler$^1$} \\
    $^1$Amazon AI Translate, Berlin \\
    $^2$University of Zurich\\ \medskip
    \texttt{\{vamvas,sennrich\}@cl.uzh.ch},
    \texttt{\{domhant,trenous,ehasler\}@amazon.com}
 }

\begin{document}
\maketitle
\begin{abstract}
    Neural metrics trained on human evaluations of MT tend to correlate well with human judgments, but their behavior is not fully understood.
    In this paper, we perform a controlled experiment and compare a baseline metric that has not been trained on human evaluations (\textit{\prism{}}) to a trained version of the same metric (\textit{\ftprism{}}).
    Surprisingly, we find that \ftprism{} becomes more robust to machine-translated references, which are a notorious problem in MT evaluation.
    This suggests that the effects of metric training go beyond the intended effect of improving overall correlation with human judgments.
\end{abstract}

\section{Introduction}
While trained evaluation metrics for machine translation~(MT) tend to have a high correlation with human judgments~\cite{freitag-etal-2022-results}, they remain black boxes, sometimes behaving in unexpected ways~\cite{amrhein-2022-identifying,rei-etal-2023-inside}.
This calls into question whether a metric's utility can be measured solely by its correlation with human judgments.

In this paper, we intentionally provide MT metrics with \textit{machine-translated reference translations}, as opposed to human-created references, and investigate how this factor influences the behavior of a metric.
In MT evaluation research, the human translators who create reference translations are usually asked to produce them from scratch, in order to avoid references that are machine-translated or post-edited~\cite{kocmi-etal-2022-findings}.
Nevertheless, traces of MT have been detected in some reference sets~\cite{kloudova-etal-2021-detecting,akhbardeh-etal-2021-findings,kocmi-etal-2022-findings}.
It is therefore important to understand how metrics behave under such references.

\begin{figure}[t!]
\centering
\includegraphics[width=\columnwidth]{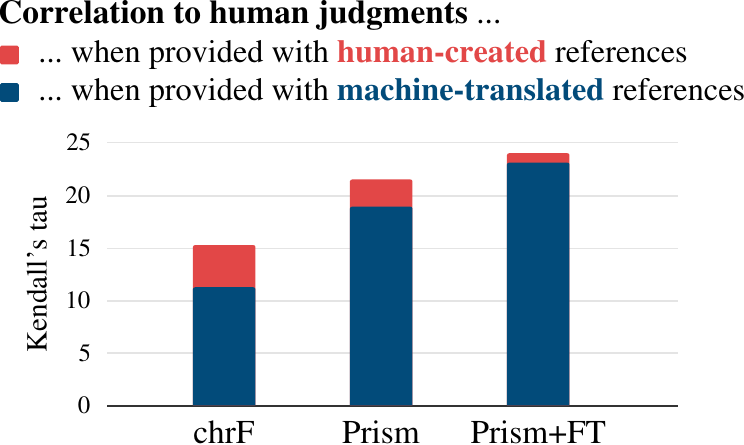}
\caption{
Metrics for MT quality have a lower segment-level correlation with human judgments when provided with machine-translated references.
However, trained metrics, such as our \ftprism{}, become more robust to the use of machine translations as references.
}
\label{fig:figure-1}
\end{figure}

In our experiments, we use a surrogate for real post-edited references in the form of error-free output by various systems from the WMT 2021 news translation task~\cite{akhbardeh-etal-2021-findings}.
Our results show that there is a stark difference between trained and non-trained metrics:
While trained metrics maintain most of their accuracy when provided with such MT-derived references, non-trained metrics exhibit a substantial drop in accuracy.

To corroborate this observation, we perform a controlled experiment involving \prism{}~\cite{thompson-post-2020-automatic}, a metric that is based on a multilingual MT system.
The original version of \prism{} can be considered non-trained, since it learns from parallel sentences without human judgments.
We then fine-tune \prism{} on a dataset of human judgments, using a bidirectional pairwise ranking approach.

As expected, the segment-level correlation of \prism{} increases during fine-tuning, indicating that the metric learns to better predict human judgments~(Figure~\ref{fig:figure-1}).
Moreover, we find that fine-tuning narrows the gap in performance between human-created and machine-translated references.
Our experiment thus indicates that training a metric on human evaluation data can influence its behavior in a way that is not captured by global correlation with human judgments.
Code to reproduce our findings will be made available.\footnote{\url{https://github.com/amazon-science/prism-finetuned}}

To summarize, the paper makes the following contributions:
\begin{itemize}
    \item We propose a metric evaluation setup that intentionally uses machine-translated references, and demonstrate that non-trained metrics perform poorly in this setup.
    \item We present an approach for fine-tuning \prism{} on human judgments that significantly improves segment-level correlation on unseen test data.
    \item We show that fine-tuning \prism{} on human judgments makes it more robust to the use of machine-translated references.
\end{itemize}

\section{Background}

\subsection{Reference-based Evaluation}
Automatic evaluation of MT is often performed by comparing the system output with one or more reference translations, using an evaluation metric.
Evaluation metrics can be roughly divided into \textit{trained} and \textit{non-trained} metrics.
Trained metrics receive supervision from human judgments of past machine translations.
For example, \citet{sellam-etal-2020-bleurt} and \citeauthor{rei-etal-2020-comet}~(\citeyear{rei-etal-2020-comet};~`COMET') fine-tuned a pre-trained sentence encoder on such human judgments, using regression or ranking objectives.

Non-trained metrics, on the other hand, rely on a heuristic to make the comparison.
Metrics such as BLEU~\cite{papieni-etal-2022-bleu} and chrF~\cite{popovic-2015-chrf} are based on the overlap of words or characters between the system output and the reference. \citet{thompson-post-2020-automatic} use the perplexity of a neural sequence-to-sequence model, called \prism{}, that has been trained on multilingual MT.
Systematic comparisons of evaluation metrics~\cite{freitag-etal-2022-results} have shown that trained metrics tend to correlate better with human judgments than non-trained metrics do, especially if the latter are based on overlap heuristics.

\subsection{Quality of Reference Translations}
The reliability of reference-based evaluation metrics also depends on the quality of the references they are provided with~\cite{freitag-etal-2021-results}.
A notorious source of noise in references is \textit{translationese}, which is characterized by monotonicity with respect to the source sequence and a high \textit{n}-gram overlap with system translations~\cite{freitag-etal-2020-bleu}.
\citet{freitag-etal-2020-bleu} have shown that translationese references cause BLEU scores to be higher, and the scores are dominated by matches of common, unspecific \textit{n}-grams.
They find that BLEU scores under non-translationese references tend to be lower, but more precise.

\citet{agrawal-etal-2023-findings} observed that post-edited references for spoken language translation seem to inflate BLEU scores, but not the scores of COMET.
However, the relationship between metric training and the quality of reference translations has not been studied in detail.
In this paper, we hypothesize that robustness to machine-translated references may partially explain why trained metrics are more accurate in practice.

\section{Experimental Setup}

\subsection{Measuring Global Correlation}
For measuring the overall correlation of a metric to human judgments, we follow the WMT 2021 metrics task~\cite{freitag-etal-2021-results} and use MQM annotations of submissions to the 2021 WMT news translation task~\cite{akhbardeh-etal-2021-findings}.
The evaluation data cover two domains, news and TED talks.
Table~\ref{tab:evaluation-data-statistics} reports statistics for these data.

We closely replicate the methodology of the WMT 2021 metrics task.
On the segment level, we report Kendall's tau coefficient across all segments and systems; on the system level, we report \textit{pairwise accuracy}~\cite{kocmi-etal-2021-ship}, i.e., the ratio of system pairs that a metric ranks in the same order as human annotators have.
Following the shared task, we only consider system translations and exclude human translations from the evaluation.
We then perform \textit{perm-both} hypothesis tests~\cite{deutsch-etal-2021-statistical} to validate metrics comparisons at $\alpha=0.05$.

\subsection{Measuring the Effect of Machine-translated References}
\begin{figure*}[t!]
\centering
\includegraphics[width=\textwidth]{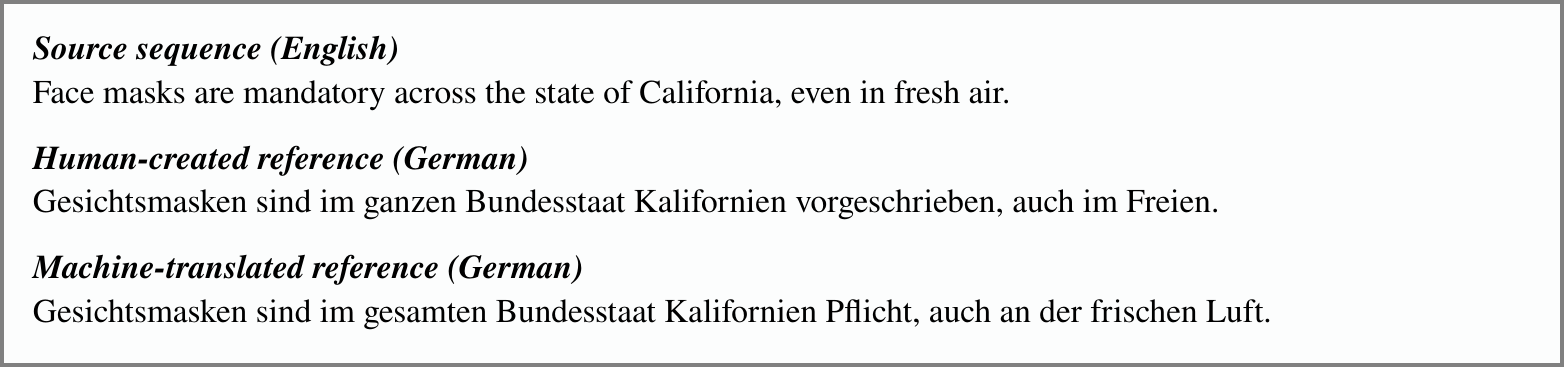}
\caption{
Example of a machine-translated reference compared to the standard reference created by a human translator. The machine-translated reference is more literal (\textit{an der frischen Luft} `in fresh air').
}
\label{fig:example}
\end{figure*}

In the context of our analysis, we use error-free system translations from the WMT 2021 news translation task as a surrogate for real post-edited references.
Specifically, we randomly select system translations that have been annotated according to the MQM standard and in which no annotator has marked an error.
This approach allows us to simulate a post-editing process without the cost and noise incurred by actual post-editing.

Figure~\ref{fig:example} and Appendix~\ref{sec:references-examples} juxtapose some examples of error-free system translations and the standard, human-created reference translations.
The former tend to be more literal and more aligned to the source, both in terms of syntax and content.

It should be noted that when we evaluate a metric in this analysis, we draw from the same set of systems and human annotations as we do for extracting the references.
We take care to properly separate the system translations used as a reference from those that are evaluated based on that reference.

\begin{figure}[t!]
\centering
\includegraphics[width=\columnwidth]{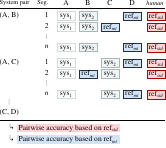}
\caption{
To measure the effect of machine-translated references, we use error-free output from other, unrelated MT systems as references.
For example, when comparing system A to system B, we use a translation from either system C, D, etc. as a reference for each segment.
}
\label{fig:evaluation-schema}
\end{figure}

To calculate segment-level correlation, we sample a random error-free translation from an unrelated system, for each system output.\footnote{Segment-level correlation is calculated jointly across all segments and systems, and as a consequence, using different references to evaluate the translations of different systems adds some noise to the correlation.
However, we expect that the correlation is dominated by the segment axis and not by the system axis.
Our findings on the segment level are consistent with our findings on the system level.}
To calculate system-level pairwise accuracy, we use different sets of references depending on the pair of systems that is compared.
Figure~\ref{fig:evaluation-schema} shows that our approach is comparable to cross-validation.
For every pair of systems that we consider when calculating the pairwise accuracy of a metric, we select one reference translation from an unrelated system, independently per segment.
As a consequence, we use slightly different reference sets for ranking different pairs of systems.

We then compare the accuracy of a metric when provided with the machine-translated references to its accuracy when using the standard references.
To ensure comparability, we skip all the segments where no machine-translated reference is available (which is either because the segment has not been part of the annotation study or because annotators have found an error in every system translation).
The metric accuracies for both $\text{ref}_{std}$ and $\text{ref}_{mt}$ are thus calculated based on a subset of the segments used to calculate global correlation.
Table~\ref{tab:evaluation-data-statistics} shows that only for one language pair a substantial number of segments need to be skipped~(Chinese--English news).
For the other language pairs, between 0\% and 4.5\% of the segments are skipped.

\section{Fine-tuning the Prism Metric}
Prism~\cite{thompson-post-2020-automatic} is a reference-based evaluation metric that relies on the paraphrasing probability between a system translation and a reference.
The probability is estimated by a multilingual NMT model as a zero-shot translation direction.
The model is expected to prefer mere copies of the source sequence to more creative paraphrases, which is especially useful for reference-based evaluation.

The NMT model uses the reference as a source sequence~$x$ and the system translation as a hypothesis~$y$, or vice versa.
The segment-level score~$S$ is then calculated from token-level log-probabilities:\footnote{This score is called~$H$ in the original definition. We use~$S$ instead, to avoid confusion with cross-entropy~(which is $-S$).}
\begin{align*}
S(y|x) &= \frac{1}{|y|} \sum_{t=1}^{|y|} \log p(y_t | y_{i<t},x).
\end{align*}
By default, Prism uses the average of both paraphrasing directions:
\[ \text{Prism}(\text{sys}, \text{ref}) = \frac{1}{2} S(\text{sys}|\text{ref}) + \frac{1}{2} S(\text{ref}|\text{sys}). \]

\noindent An overall score for a system can then be calculated as an average over a collection of segments.

\subsection{Training Objective}
In order to fine-tune Prism, we combine a standard cross-entropy objective and a bidirectional pairwise ranking objective.

For the \textit{cross-entropy objective}, we use the source sequence ($\text{src}$) and the reference translation ($\text{ref}$) of the training examples to continue the cross-entropy training:
\[ L_{\text{src}\rightarrow{}\text{ref}} = - S(\text{ref}|\text{src}). \]
Our goal in using this objective is to familiarize Prism with the segments to which the human judgments refer, and to prevent catastrophic forgetting during the fine-tuning stage.

In addition, we propose a \textit{bidirectional pairwise ranking objective}.
In the forward direction, we train Prism to correctly rank two system translations ($\text{sys}^+$ and $\text{sys}^-$), conditioned on the reference~(\textit{forward ranking}):
\begin{align*}
L_{\text{ref}\rightarrow{}\text{sys}} = \max\{ 0, \epsilon &- S(\text{sys}^+|\text{ref}) \\
                                                                 &+ S(\text{sys}^-|\text{ref})\},
\end{align*}
where $\epsilon$ is a margin value.
We add a second ranking loss for the reverse paraphrasing direction, i.e., for reconstructing the reference from either of the system translations~(\textit{backward ranking}):
\begin{align*}
L_{\text{sys}\rightarrow{}\text{ref}} = \max\{ 0, \epsilon &- S(\text{ref}|\text{sys}^+) \\
                                                                 &+ S(\text{ref}|\text{sys}^-)\}.
\end{align*}
The complete fine-tuning objective is:
\[ L = \alpha L_{\text{src}\rightarrow{}\text{ref}} + (\frac{1}{2} L_{\text{ref}\rightarrow{}\text{sys}} + \frac{1}{2} L_{\text{sys}\rightarrow{}\text{ref}}), \]
where $\alpha$ is a scalar to balance the two terms.

\begin{figure}[]
\centering
\includegraphics[width=\columnwidth]{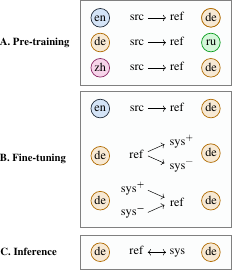}
\caption{
Schematic illustration of the sequences used for pre-training, fine-tuning, and applying the Prism model to MT evaluation.
Prism has been~\textbf{(A)} pre-trained on multilingual translation to and from~39 languages as described by \citet{thompson-post-2020-automatic}; inference~\textbf{(C)} makes use of the zero-shot paraphrasing capability acquired by the model during pre-training.
We add a fine-tuning stage~\textbf{(B)} with data derived from human evaluations of MT.
In this illustration, Prism is fine-tuned on English--German examples.
}
\label{fig:prism-schema}
\end{figure}

Figure~\ref{fig:prism-schema} is a schematic illustration of the objectives for pre-training, fine-tuning, and inference.

\subsection{Training Data}
For fine-tuning Prism, we use human judgments of submissions to the 2020 WMT news translation tasks~\cite{barrault-etal-2020-findings}, collected by \citet{freitag-etal-2021-experts}.\footnote{Submission data are available at \url{https://github.com/google-research/mt-metrics-eval} and the MQM annotations are available at \url{https://github.com/google/wmt-mqm-human-evaluation}}
These annotations are based on the Multidimensional Quality Metrics~(MQM) framework~\cite{lommel-etal-2014-mqm} and have been shown to correlate better with automatic metrics than previous direct assessments, especially when the evaluation concerns high-quality translations~\cite{freitag-etal-2021-experts,freitag-etal-2021-results}.
Specifically, we train Prism on human judgments for English--German and Chinese--English translations of news.
We train a single model jointly on both language pairs.

To use the human judgments for training on pairwise ranking, the direct MQM assessments need to be converted into relative rankings of translation pairs.
In previous work, direct~(non-MQM) assessments have been normalized and aggregated across annotators before being compared~\cite{ma-etal-2019-results}.
Since MQM ratings are known to have low inter-annotator agreement on the segment level~\cite{freitag-etal-2021-results}, we opt for intra-annotator pairing instead.
Specifically, we only pair translations that have been rated by the same annotator, and we do not compare MQM scores across annotators.
Relative rankings are created independently for each annotator and then concatenated.
Furthermore, we only pair translations that have a score difference greater than 0.1, which would correspond to a minor fluency or punctuation error.
Taken together, these criteria should ensure there is a noticeable difference between the quality of two system translations $\text{sys}^+$ and $\text{sys}^-$ in the eyes of at least one annotator.
We hold out 5000 relative rankings from the resulting training data as a validation set and use it to select hyperparameters.
Detailed statistics for the training data are provided in Table~\ref{tab:training-data-statistics}.

\subsection{Implementation Details}
The fine-tuning was implemented in Fairseq~\cite{ott-etal-2019-fairseq}.
We start with the original \texttt{Prism39} model released by \citet{thompson-post-2020-automatic}.\footnote{\url{https://data.statmt.org/prism/}}
We then fine-tune the model for a single pass over the training data, using Adam.
The initial learning rate is set to 1e-4 without any warm-up steps.
We use half-precision training and an effective batch size of 360k tokens.
Other settings match the pre-training setup of Prism.

We set the margin hyperparameter $\epsilon$ to 0.1, and the cross-entropy weight $\alpha$ to 0.1 as well.
The hyperparameters have been selected based on segment-level correlation on the validation set.
Since we jointly train on two language pairs, we iterate over batches for each language pair in a round-robin fashion, upsampling the smaller language pair.
Fine-tuning takes about one hour on a \texttt{p3.8xlarge} AWS instance, which has 4 Tesla V100 GPUs with 16 GB of memory.

\section{Results}

\begin{table}[htb!]
\begin{tabularx}{\columnwidth}{@{}Xrrr@{}}
\toprule
& \textsc{en--de} & \textsc{en--ru} & \textsc{zh--en} \\ \midrule
\prism{} & 19.3                & 22.4       & 28.8      \\
\ftprism{}                        & \textbf{25.3} & \textbf{23.7}    & \textbf{31.5}  \\ \bottomrule
\end{tabularx}
\caption{
In-domain accuracy of Prism on WMT 2021 news translation submissions.
We report segment-level Kendall's tau correlation to human judgments.
Bold font denotes that the improvement achieved through fine-tuning is significant with $\alpha=0.05$.
Note that \ftprism{} has not been fine-tuned on the \textsc{en--ru} language pair.
}
\label{tab:results-wmt21-news}
\end{table}

\begin{table}[htb!]
\begin{tabularx}{\columnwidth}{@{}Xrrr@{}}
\toprule
& \textsc{en--de} & \textsc{en--ru} & \textsc{zh--en} \\ \midrule
\prism{} & 24.2                & 21.9       & 19.6      \\
\ftprism{}                        & \textbf{26.9} & 22.3    & \textbf{21.9}  \\ \bottomrule
\end{tabularx}
\caption{
Out-of-domain accuracy of Prism on WMT 2021 system translations of TED talks in terms of segment-level Kendall's tau.
Bold indicates that the improvement is significant with $\alpha=0.05$.
}
\label{tab:results-wmt21-tedtalks}
\end{table}

\begin{table*}[]
\begin{tabularx}{\textwidth}{@{}Xrrrrrrrr@{}}
\toprule
                    & \multicolumn{2}{l}{\textsc{en--de}} & \multicolumn{2}{l}{\textsc{en--ru}} & \multicolumn{2}{l}{\textsc{zh--en}} &
                      \multicolumn{2}{l}{Average} \\
                    & $\text{ref}_{std}$            & $\text{ref}_{mt}$\phantom{mmm.\,\,} & $\text{ref}_{std}$            & $\text{ref}_{mt}$\phantom{mmm.\,\,} & $\text{ref}_{std}$            & $\text{ref}_{mt}$\phantom{mmm.\,\,} & $\text{ref}_{std}$            & $\text{ref}_{mt}$\phantom{mmm.\,\,} \\ \midrule
BLEU & 8.4 & 7.0\dr{-16.7} & 12.1 & 11.8\drs{-2.5} & 15.2 & 14.8\drs{-2.6} & 11.9 & 11.2\drs{-5.9} \\
chrF & 11.1 & 8.3\dr{-25.2} & 19.3 & 13.8\dr{-28.5} & 16.7 & 15.7\drs{-6.0} & 15.7 & 12.6\dr{-19.7} \\
Prism & 18.9 & 18.2\drs{-3.7} & 22.4 & 20.6\drs{-8.0} & 24.2 & 23.5\drs{-2.9} & 21.8 & 20.8\drs{-4.9} \\ \midrule
Prism+FT & 24.9 & 24.4\drs{-2.0} & 23.7 & 22.3\drs{-5.9} & 26.6 & 26.8\dgsp{0.8} & 25.1 & 24.5\drs{-2.3} \\
COMET & 25.1 & 24.6\drs{-2.0} & 27.6 & 25.4\drs{-8.0} & 32.1 & 32.1\dgsp{0.0} & 28.3 & 27.4\drs{-3.2} \\ \bottomrule
\end{tabularx}
\caption{Segment-level correlation of MT metrics when provided with the standard references~($\text{ref}_{std}$) of the WMT21 metrics news subtask~\cite{freitag-etal-2021-results}, and with machine-translated references~($\text{ref}_{mt}$).
The percentages denote the relative change in correlation when falling back to machine-translated references.
The trained metrics, \ftprism{} and COMET~(\texttt{wmt21-comet-mqm}), have a more favorable relative change than the non-trained metrics, which indicates higher robustness to machine-translated references.}
\label{tab:results-segment}
\end{table*}

\begin{table*}[htb!]
\begin{tabularx}{\textwidth}{@{}Xrrrrrrrr@{}}
\toprule
                    & \multicolumn{2}{l}{\textsc{en--de}} & \multicolumn{2}{l}{\textsc{en--ru}} & \multicolumn{2}{l}{\textsc{zh--en}} &
                      \multicolumn{2}{l}{Average} \\
                    & $\text{ref}_{std}$            & $\text{ref}_{mt}$\phantom{mmm.\,\,} & $\text{ref}_{std}$            & $\text{ref}_{mt}$\phantom{mmm.\,\,} & $\text{ref}_{std}$            & $\text{ref}_{mt}$\phantom{mmm.\,\,} & $\text{ref}_{std}$            & $\text{ref}_{mt}$\phantom{mmm.\,\,} \\ \midrule
BLEU & 89.7 & 74.4\dr{-17.1} & 70.3 & 58.2\dr{-17.2} & 61.5 & 61.5\dgsp{0.0} & 73.8 & 64.7\dr{-12.4} \\
chrF & 87.2 & 71.8\dr{-17.7} & 74.7 & 56.0\dr{-25.0} & 60.3 & 56.4\drs{-6.5} & 74.1 & 61.4\dr{-17.1} \\
Prism & 85.9 & 73.1\dr{-14.9} & 83.5 & 62.6\dr{-25.0} & 61.5 & 56.4\drs{-8.3} & 77.0 & 64.0\dr{-16.8} \\ \midrule
Prism+FT & 89.7 & 80.8\drs{-9.9} & 80.2 & 61.5\dr{-23.3} & 61.5 & 61.5\dgsp{0.0} & 77.1 & 67.9\dr{-11.9} \\
COMET & 79.5 & 84.6\dgsp{6.4} & 68.1 & 65.9\drs{-3.2} & 60.3 & 55.1\drs{-8.6} & 69.3 & 68.5\drs{-1.1} \\ \bottomrule
\end{tabularx}
\caption{System-level pairwise accuracy of MT metrics when provided with the standard references of the WMT21 metrics news subtask~\cite{freitag-etal-2021-results}, and with machine-translated references. Again, the trained metrics, \ftprism{} and COMET (\texttt{wmt21-comet-mqm}), tend to be more robust to machine-translated references.
}
\label{tab:results-system}
\end{table*}

\paragraph{Effect of fine-tuning Prism}
Table~\ref{tab:results-wmt21-news} shows that fine-tuning Prism has the intended effect:
\textit{Fine-tuning Prism on human judgments of machine translations significantly improves correlation with human judgments on an unseen test set.}
The effect of fine-tuning is especially pronounced for the English--German and Chinese--English language pairs, since the metric was fine-tuned on those pairs.
Interestingly, we also observe positive cross-lingual transfer to the English--Russian language pair, which was not seen during fine-tuning.
Table~\ref{tab:results-wmt21-tedtalks} shows that the positive effect of fine-tuning extends to the TED Talks domain, even though the metric was not fine-tuned on this domain.

\paragraph{Effect of using machine-translated references}
Table~\ref{tab:results-segment} reports the segment-level correlation of different metrics when using either standard references or machine-translated references.
Note that the values for \prism{} slightly differ from Tables~\ref{tab:results-wmt21-news} and~\ref{tab:results-wmt21-tedtalks} because this analysis is based on a subset of the segments.
\textit{We find that the correlation of metrics to human judgments tends to decrease under machine-translated references.}
For the Chinese--English dataset the relative decline is smaller than average, but is still noticeable for most metrics.

In Table~\ref{tab:results-segment}, when comparing the non-trained metrics (above the horizontal line) to the trained metrics (below the line), we observe that the decline in correlation is smaller for the trained metrics.
An especially interesting comparison is between \prism{} and \ftprism{}, given that the two metrics differ only in the training data.
\textit{\ftprism{} is consistently more robust to machine-translated references than \prism{}, indicating that the metric learns to cope with such references during the fine-tuning stage.}

With respect to system-level pairwise accuracy~(Table~\ref{tab:results-system}), we observe a similar trend.
\ftprism{} does not show significantly higher pairwise accuracy than \prism{} when using standard references, which is explained by the high statistical variance of the pairwise accuracy metric.
But again, \ftprism{} appears more robust to machine-translated references than \prism{}.
Finally, Appendix~\ref{appendix:ted-talks} reports results for the TED talks domain, where the same patterns can be observed.

\paragraph{Ablation Study}
We perform an ablation study to measure the influence to the three terms in the Prism fine-tuning objective.
Appendix~\ref{appendix:ablation} shows that removing either of the three terms decreases segment-level correlation.
The ablation shows that the cross-entropy objective has the additional effect of stabilizing the model:
Without cross-entropy, the average probability scores output by Prism shift from 0.47 to 0.35 after a single epoch of fine-tuning, and the BLEU achieved by the Prism translation model on an unseen test set clearly declines.

\section{Related Work}

\paragraph{Machine translations as references}
\citet{popovic-etal-2016-potential} first investigated the potential of using post-edited machine translations as references, finding that post-edited translations stemming from high-quality systems are better references than those from low-quality systems.
\citet{toral-2019-post} argued that post-edited machine translations can be seen as an exacerbated form of translationese (\textit{post-editese}).
Combined with the finding of \citet{freitag-etal-2020-bleu} that translationese references are less favorable than intentionally paraphrased references, this suggests that machine translations, even if post-edited, are a challenge for MT evaluation.

\citet{albrecht-hwa-2007-regression} propose to train an evaluation metric using non-annotated translations of other systems as \textit{pseudo-references}.
They hypothesize that a metric can learn to detect and to constructively utilize any errors in these references.
\citet{yoshimura-etal-2019-filtering} instead use a paraphrase identifier to filter pseudo-references based on their paraphrastic similarity to a human-created reference.
Finally, minimum Bayes risk decoding~\cite{kumar-byrne-2004-minimum} employs pseudo-references for generating translations, and has been shown to depend on robust metrics as well~\cite{freitag-etal-2022-mbr,amrhein-2022-identifying}.

\paragraph{Training a sequence-to-sequence model on pairwise ranking}
Pairwise ranking has commonly been used to train SVM~\cite{ye-etal-2007-sentence,duh-2008-ranking,stanojevic-simaan-2014-fitting} and neural network encoders~\cite{guzman-etal-2015-pairwise,dusek-etal-2019-automatic}.
A more recent approach has been to fine-tune pre-trained sentence encoders so that the embedding similarities of two hypotheses and the reference and/or source are optimized for pairwise ranking~\cite{rei-etal-2020-comet,zhang-van-genabith-2020-translation}, in which case the max-margin loss reduces to a triplet margin loss~\cite{schroff-etal-2015-facenet}.
In this paper, we do not rely on the similarity of sentence embeddings but use the perplexity of a sequence-to-sequence model as a metric.

Since we optimize perplexity given positive and negative examples, our fine-tuning approach becomes very similar to contrastive learning for NMT.
Typical applications of contrastive learning try to eliminate specific translation error types by creating perturbed versions of the training references~\cite{yang-etal-2019-reducing,hwang-etal-2021-contrastive}.
A similar objective has been used for discriminative re-ranking of translation candidates~\cite{shen-etal-2004-discriminative,yu-etal-2020-deepmind}.
In this paper, however, the goal is not to improve translation output but to train an evaluation metric on human judgments.

\section{Conclusion}
We have shown that metrics without supervision by human judgments, such as BLEU and chrF, tend to be inaccurate under machine-translated references, while trained metrics are more robust.
In order to methodically examine this phenomenon, we have trained the Prism evaluation metric on a dataset of human judgments.
Our experiments show that fine-tuning improves the segment-level accuracy of Prism on an unseen test set across multiple language pairs and domains, and clearly increases its robustness to machine-translated references.

One conclusion to draw from our findings is that post-edited references likely diminish the accuracy of reference-based metrics and should be avoided.
A second conclusion is that if it cannot be ruled out that references originate from MT, as is often the case in practice, trained metrics are to be preferred.
Fine-tuning a metric such as Prism on reference-based evaluation can thus be seen as a technique to let the metric make the best out of reference translations in the wild.

\section*{Limitations}
Our study is mainly limited by the data we use for fine-tuning and evaluating Prism.
The experiments are based on three language pairs only.
Automatic MT evaluation is relevant for many more language pairs and language families, including and maybe especially so for low-resource settings.

Secondly, it should be mentioned that the machine translations we use in our analysis have been generated by systems based on a similar technology.
Almost all of the systems seem to use the Transformer architecture, and they have all been trained on similar data~\cite{akhbardeh-etal-2021-findings}.
It is possible that our findings do not generalize to the evaluation of other varieties of MT, such as rule-based systems, or to reference-based evaluation metrics that use large language models~\cite{kocmi-federmann-2023-large}.

\section*{Acknowledgements}
We thank Bill Byrne, Felix Hieber, Brian Thompson and Ke Tran for comments on an earlier stage of this project.
JV and RS acknowledge funding by the Swiss National Science Foundation (project MUTAMUR; no. 176727).

\bibliography{bibliography}
\bibliographystyle{acl_natbib}

\appendix

\setcounter{table}{0}
\renewcommand{\thetable}{A\arabic{table}}

\onecolumn

\section{Ablation Study}
\label{appendix:ablation}

\begin{table*}[hbt!]
\begin{tabularx}{\textwidth}{@{}Xrrr@{\hskip 20pt}rr@{}}
\toprule
\multirow{2}{*}{Variant} &
  \multicolumn{1}{r}{\multirowcell{2}[0pt][r]{Segment-level\\Kendall's tau}} &
  \multicolumn{1}{r}{\multirowcell{2}[0pt][r]{Pairwise\\accuracy}} &
  \multicolumn{1}{r@{\hskip 20pt}}{\multirowcell{2}[0pt][r]{Magnitude\\of scores}} &
  \multicolumn{2}{r@{}}{BLEU (newstest21)} \\ \cmidrule(l){5-6}
 &
  \multicolumn{1}{l}{} &
  \multicolumn{1}{l}{} &
  \multicolumn{1}{c}{} &
  \multicolumn{1}{@{\hskip 20pt}r}{\textsc{en--de}} &
  \multicolumn{1}{r@{}}{\textsc{zh--en}} \\ \midrule
Prism (no fine-tuning)                                              & 23.5 & 78.7 & 0.47 & 25.6 & 18.7 \\ \midrule
\mbox{\ftprism{}}        & 26.8 & 76.7 & 0.37 & 23.0 & 21.0 \\
– without cross-entropy                                             & 26.6 & 74.6 & 0.35 & 10.2 & 9.6 \\
– without forward ranking                                           & 26.0 & 79.2 & 0.40 & 21.9 & 20.1 \\
– without backward ranking                                          & 25.6 & 77.7 & 0.39 & 21.1 & 20.3 \\ \bottomrule
\end{tabularx}
\caption{
Ablation study for the proposed fine-tuning objective, based on the in-domain meta-evaluation setting~(WMT 2021 news translations).
In every row we remove one aspect of the fine-tuning setup.
Meta-metrics are averaged across three language pairs.
\textit{Magnitude of scores} refers to the average segment-level scores predicted by the Prism model, converted to probability space via $2^x$.
}
\label{tab:results-ablation}
\end{table*}

\vfill

\section{Evaluation on TED Talks}\label{appendix:ted-talks}

\begin{table*}[htb!]
\begin{tabularx}{\textwidth}{@{}Xrrrrrrrr@{}}
\toprule
                    & \multicolumn{2}{l}{\textsc{en--de}} & \multicolumn{2}{l}{\textsc{en--ru}} & \multicolumn{2}{l}{\textsc{zh--en}} &
                      \multicolumn{2}{l}{Average} \\
                    & $\text{ref}_{std}$            & $\text{ref}_{mt}$\phantom{mmm.\,\,} & $\text{ref}_{std}$            & $\text{ref}_{mt}$\phantom{mmm.\,\,} & $\text{ref}_{std}$            & $\text{ref}_{mt}$\phantom{mmm.\,\,} & $\text{ref}_{std}$            & $\text{ref}_{mt}$\phantom{mmm.\,\,} \\ \midrule
BLEU & 13.4 & 7.1\dr{-47.0} & 16.0 & 12.8\dr{-20.0} & 11.0 & 9.1\dr{-17.3} & 13.5 & 9.7\dr{-28.2} \\
chrF & 14.3 & 7.9\dr{-44.8} & 18.9 & 12.8\dr{-32.3} & 11.4 & 9.0\dr{-21.1} & 14.9 & 9.9\dr{-33.4} \\
Prism & 23.6 & 17.7\dr{-25.0} & 22.0 & 17.5\dr{-20.5} & 18.0 & 15.9\dr{-11.7} & 21.2 & 17.0\dr{-19.7} \\ \midrule
Prism+FT & 26.4 & 24.2\drs{-8.3} & 22.2 & 21.6\drs{-2.7} & 20.2 & 19.4\drs{-4.0} & 22.9 & 21.7\drs{-5.2} \\
COMET & 27.3 & 24.6\drs{-9.9} & 25.8 & 23.2\dr{-10.1} & 20.8 & 20.7\drs{-0.5} & 24.6 & 22.8\drs{-7.3} \\ \bottomrule
\end{tabularx}
\caption{Segment-level correlation of MT metrics when provided with the standard references and with machine-translated references. The percentages denote the relative change in correlation when falling back to machine-translated references.}
\label{tab:results-tedtalks}
\end{table*}
\begin{table*}[htb!]
\begin{tabularx}{\textwidth}{@{}Xrrrrrrrr@{}}
\toprule
                    & \multicolumn{2}{l}{\textsc{en--de}} & \multicolumn{2}{l}{\textsc{en--ru}} & \multicolumn{2}{l}{\textsc{zh--en}} &
                      \multicolumn{2}{l}{Average} \\
                    & $\text{ref}_{std}$            & $\text{ref}_{mt}$\phantom{mmm.\,\,} & $\text{ref}_{std}$            & $\text{ref}_{mt}$\phantom{mmm.\,\,} & $\text{ref}_{std}$            & $\text{ref}_{mt}$\phantom{mmm.\,\,} & $\text{ref}_{std}$            & $\text{ref}_{mt}$\phantom{mmm.\,\,} \\ \midrule
BLEU & 66.7 & 35.9\dr{-46.2} & 83.5 & 58.2\dr{-30.3} & 64.1 & 65.4\dgsp{2.0} & 71.4 & 53.2\dr{-25.6} \\
chrF & 65.4 & 46.2\dr{-29.4} & 85.7 & 53.8\dr{-37.2} & 61.5 & 66.7\dgsp{8.5} & 70.9 & 55.6\dr{-21.6} \\
Prism & 69.2 & 44.9\dr{-35.1} & 82.4 & 48.4\dr{-41.3} & 67.9 & 66.7\drs{-1.8} & 73.2 & 53.3\dr{-27.1} \\ \midrule
Prism+FT & 66.7 & 51.3\dr{-23.1} & 81.3 & 61.5\dr{-24.4} & 62.8 & 70.5\dgp{12.3} & 70.3 & 61.1\dr{-13.0} \\
COMET & 84.6 & 53.8\dr{-36.4} & 78.0 & 74.7\drs{-4.2} & 67.9 & 75.6\dgp{11.3} & 76.8 & 68.0\dr{-11.5} \\ \bottomrule
\end{tabularx}
\caption{System-level pairwise accuracy of MT metrics when provided with the standard references and with machine-translated references.}
\label{tab:results-tedtalks-system}
\end{table*}

\vfill

\pagebreak

\section{Training Data Statistics}
\begin{table*}[htb!]
\begin{tabularx}{\textwidth}{@{}Xr@{\hskip 20pt}r@{}}
\toprule
\textbf{Language pair}                            & \multicolumn{1}{r@{\hskip 20pt}}{\textbf{\textsc{en--de}}} & \multicolumn{1}{r@{}}{\textbf{\textsc{zh--en}}} \\ \midrule
Number of systems (including sets of human translations)                        & 10                        & 10                        \\ \midrule
Number of annotated segments             & 1\,418                      & 2\,000                      \\
– used for relative rankings             & 1\,411                      & 1\,985                      \\ \midrule
Number of annotated system translations & 14\,110                     & 19\,994                     \\
– used for relative rankings             & 14\,110                     & 19\,850                     \\ \midrule
Number of relative rankings              & 126\,217                    & 164\,137                    \\
– training split                         & 121\,217                    & 159\,137                    \\
– validation split                       & 5\,000                      & 5\,000                      \\ \bottomrule
\end{tabularx}
\caption{Statistics for the WMT 2020 MQM ratings~\cite{freitag-etal-2021-experts} and for the relative rankings that we derive using an intra-annotator pairing approach.}
\label{tab:training-data-statistics}
\end{table*}

\vfill

\section{Meta-Evaluation Data Statistics}
\begin{table*}[htb!]
\begin{tabularx}{\textwidth}{@{}Xrrrrrr@{}}
\toprule
                                                    & \multicolumn{3}{c}{News} & \multicolumn{3}{c}{TED Talks}  \\  \cmidrule(l){2-4} \cmidrule(l){5-7}
                                                    & \textsc{en--de}  & \textsc{en--ru}  & \textsc{zh--en}  & \textsc{en--de}    & \textsc{en--ru}    & \textsc{zh--en}   \\ \midrule
Number of systems (without human)                   & 13     & 14     & 13     & 13       & 14       & 13      \\ \midrule
Number of MQM-annotated segments                    & 527    & 527    & 650    & 529      & 512      & 529     \\ \midrule
Number of segments with machine-translated \\reference (on average across system pairs) & 518 & 527 & 461 & 517 & 511 & 505 \\ \bottomrule
\end{tabularx}
\caption{Statistics for the WMT 2021 MQM ratings~\cite{freitag-etal-2021-results} we use for evaluating the metrics.
}
\label{tab:evaluation-data-statistics}
\end{table*}

\vfill

\section{Model Hyperparameters}\label{sec:model-hyperparameters}
\begin{table*}[htb!]
\begin{tabularx}{\textwidth}{@{}Xrrrrrr@{}}
\toprule
Model  &
$N$ &
$d_{\text{model}}$ &
$d_{\text{ffn}}$  &
$h$  &
Parameters &
Vocabulary size \\ \midrule
Prism~\cite{thompson-post-2020-automatic}                            & 16 & 1280 & 12288 & 20 & 745M  & 64k \\
\texttt{wmt21-comet-mqm}~\cite{rei-etal-2021-references}              & 24 & 1024 & 4096  & 16 & 581M  & 250k \\ \bottomrule
\end{tabularx}
\caption{Hyperparameters of the Transformer-based metrics.}
\label{tab:model-hyperparameters}
\end{table*}

\vfill
\vfill

\pagebreak

\section{Additional Examples of Human-created and Machine-translated References}\label{sec:references-examples}
\bigskip

\subsection*{English--German News Example}
\textit{Source sequence:} \\ Face masks are mandatory across the state of California, even in fresh air.

\smallskip
\noindent\textit{Standard reference:} \\ Gesichtsmasken sind im ganzen Bundesstaat Kalifornien vorgeschrieben, auch im Freien.

\smallskip
\noindent\textit{Randomly sampled error-free system translation (Nemo):} \\ Gesichtsmasken sind im gesamten Bundesstaat Kalifornien Pflicht, auch an der frischen Luft.

\bigskip
\subsection*{Chinese--English News Example}
\textit{Source sequence:} \\ \begin{CJK*}{UTF8}{gbsn}他已承认，是自己在教堂里点火。\end{CJK*}

\smallskip
\noindent\textit{Standard reference:} \\ The parish volunteer has admitted that he had started the fire in the church.

\smallskip
\noindent\textit{Randomly sampled error-free system translation (metricsystem5):} \\ He has admitted that it was himself who set the fire in the church.

\bigskip
\subsection*{English--German TED Talks Example}
\textit{Source sequence:} \\ Today I'd like to show you the future of the way we make things.

\smallskip
\noindent\textit{Standard reference:} \\ Ich möchte Ihnen heute zeigen, wie wir in Zukunft Dinge herstellen werden.

\smallskip
\noindent\textit{Randomly sampled error-free system translation (Online-W):} \\ Heute möchte ich Ihnen die Zukunft der Art und Weise zeigen, wie wir Dinge herstellen.

\bigskip
\subsection*{Chinese--English TED Talks Example}
\textit{Source sequence:} \\ \begin{CJK*}{UTF8}{gbsn}今天我想向各位展示 未来我们制作东西的方式。\end{CJK*}

\smallskip
\noindent\textit{Standard reference:} \\ Today I'd like to show you the ways we make things in the future.

\smallskip
\noindent\textit{Randomly sampled error-free system translation (metricsystem1):} \\ Today I want to show you how we will make things in the future.

\end{document}